\documentclass[10pt,twocolumn,letterpaper]{article}

\usepackage{icb}
\usepackage{times}
\usepackage{epsfig}
\usepackage{graphicx}
\usepackage{amsmath}
\usepackage{amssymb}
\usepackage{epigraph}
\usepackage{enumitem}
\usepackage{balance}
\usepackage{csquotes}
\usepackage{caption}
\usepackage{subcaption}
\usepackage{tablefootnote}
\graphicspath{ {Figures/} }
\usepackage{multirow}
\usepackage[dvipsnames]{xcolor}

\usepackage{fancyhdr}
\usepackage{xcolor}
\usepackage{xwatermark}
\newcounter{custom}
\renewcommand*{\thecustom}{%
  \textbf{%
    \arabic{custom}%
  }%
}

\usepackage[pagebackref=true,breaklinks=true,letterpaper=true,colorlinks,bookmarks=false]{hyperref}

\icbfinalcopy 

\ificbfinal\fi

\fancypagestyle{plain}{
	\fancyhf{} 
	\fancyhead[C]{\textcolor{red}{Arun Ross et al., ``Some Research Problems in Biometrics: The Future Beckons'',\\
	Proc. of 12th IAPR International Conference on Biometrics (ICB), (Crete, Greece), June 2019.}}
	\fancyfoot[R]{}

}

\begin{document}

\title{Some Research Problems in Biometrics: The Future Beckons}

\author{Arun Ross\thanks{Email: rossarun@cse.msu.edu. Web: http://iprobe.cse.msu.edu \hspace{1cm} 978-1-7281-3640-0/19/\$31.00 \copyright 2019 IEEE}\\
Sudipta Banerjee, Cunjian Chen, Anurag Chowdhury, Vahid Mirjalili,\\Renu Sharma, Thomas Swearingen, Shivangi Yadav\\
{\em Michigan State University, East Lansing, MI 48824 USA} \\
}

\maketitle
\setcounter{footnote}{0}


\begin{abstract}
The need for reliably determining the identity of a person is critical in a number of different domains ranging from personal smartphones to border security; from autonomous vehicles to e-voting; from tracking child vaccinations to preventing human trafficking; from crime scene investigation to personalization of customer service. Biometrics, which entails the use of biological attributes such as face, fingerprints and voice for recognizing a person, is being increasingly used in several such applications. While biometric technology has made rapid strides over the past decade, there are several fundamental issues that are yet to be satisfactorily resolved. In this article, we will discuss some of these issues and enumerate some of the exciting challenges in this field.    
\end{abstract}


\section{Introduction}
Biometrics refers to the automated or semi-automated recognition of individuals based on their physical, behavioral or psychophysiological traits~\cite{Jain2011Book}. These traits include face, fingerprints, iris (physical); gait, keyboard typing pattern, signature (behavioral); ECG, EEG, and saccadic eye movement (psychophysiological). A classical biometric system may be viewed as a pattern recognition engine that extracts a set of discriminative features from the input biometric data and compares this against a set of stored ``templates" in order to determine a match. Thus, a significant number of early papers in the biometric literature dealt with data acquisition, quality enhancement, feature extraction, and matching. However, the study of biometrics extends beyond pattern recognition and engages researchers from many different fields such as computer vision, signal processing, cognitive psychology, sensor design, forensics, information security, physiology, genetics, human factors, cryptography, jurisprudence, ethics, \etc. Further, since a biometric system deals with the personal information of an individual, aspects related to data privacy are also being addressed. Thus, an operational biometric system has to contend with a broad gamut of problems ranging from robust pattern recognition to provable data security/privacy in diverse scenarios.

\begin{figure*}[htpb]		
	\centering
	\includegraphics[scale = 0.5]{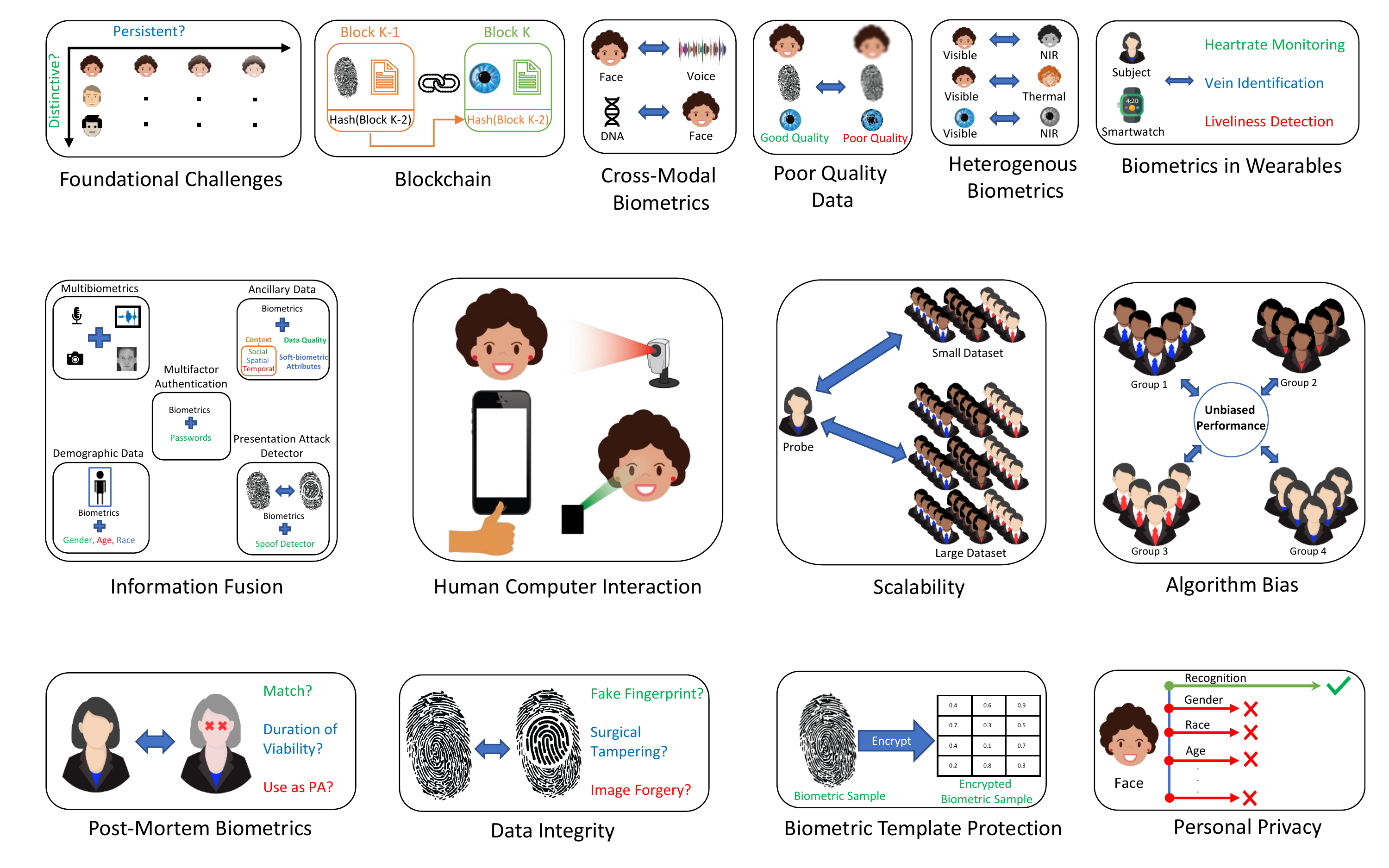}
	\caption{Illustration of some research problems in the field of biometrics.}
	\label{fig:overview}
\end{figure*}

The past decade has witnessed significant technical progress in the field of biometrics~\cite{Jain2016-50years}. This includes: (a) incorporation of compact biometric sensors in small personal devices like smartphones; (b) deployment of biometric systems in large-scale identification and de-duplication applications, such as India's Unique ID program that uses face, fingerprint and iris; (c) development of robust matching techniques for various biometric modalities based on Deep Learning; (d) investigation of previously under-explored biometric traits (such as ECG) for use in wearable devices; (e) methods for rapidly searching through large biometric databases; and (f) design of countermeasures for addressing various types of adversarial attacks against biometric systems. Notwithstanding this progress, there are a number of fundamental problems that are yet to be resolved in the field of biometrics. In this article, we highlight a few of these challenges and discuss the research opportunities in the field (Figure \ref{fig:overview}).


\section{Fundamental Science}
Biometric recognition is based on two central tenets: {\em distinctiveness} and {\em persistence} of the biometric trait of an individual. Distinctiveness is a measure of the uniqueness of a biometric trait to an individual and indicates how that biometric trait varies across the population. Persistence, on the other hand, is a measure of the temporal stability of the biometric trait pertaining to an individual. Surprisingly, our knowledge about the distinctiveness and persistence of even the four most extensively studied biometric traits (fingerprint, face, iris and voice) is incomplete~\cite{Pankanti02Individuality, DaugmanProbingUniqueIris06,Yoon2015} and often relegated to anecdotal interpretation of error rates rather than a systematic exploration of the biology of the trait~\cite{StrengtheningFS2009}.

{\bf \underline{Research problem \refstepcounter{custom}\thecustom}: Designing robust models for quantifying the uniqueness and permanence of a biometric trait.}

\section{Sensor and Human-Computer Interface}


Almost every biometric system either implicitly or explicitly imposes some type of constraint on the user or the environment during data acquisition. As an example, an iris recognition system might expect the user to position their face in a certain way with respect to the camera; similarly, a speaker recognition system might require the environment to be reasonably quiet. For ``ubiquitous biometric recognition" to gain traction, such constraints have to be surmounted in order to seamlessly recognize individuals, \ie, the interaction between an individual and a biometric system should be transparent. This would necessitate the design of novel sensors, innovative human computer interfaces and robust data processing algorithms.

\begin{figure}[tbph]		
	\centering
	\includegraphics[scale=0.75]{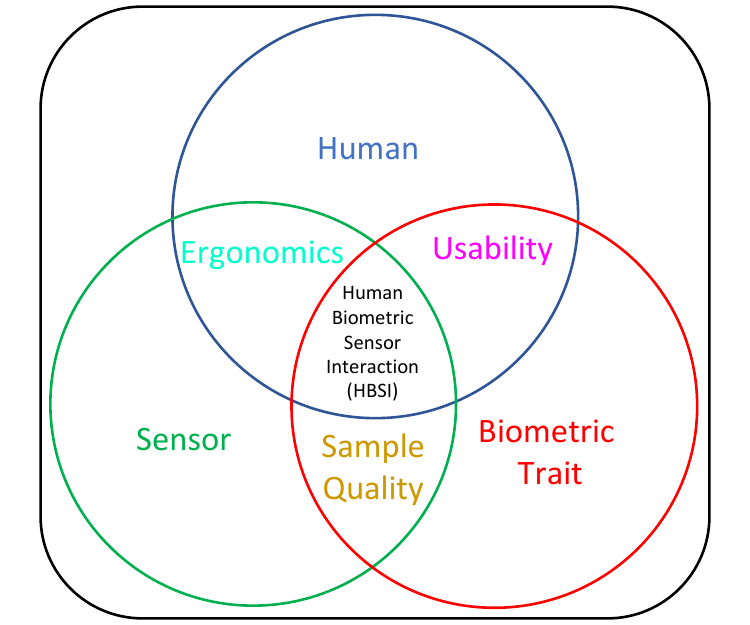}
	\caption{The design of the human-computer interface of a biometric system plays a crucial role in facilitating seamless interaction between a person and the biometric sensor in both cooperative and non-cooperative scenarios. Adapted from ~\cite{Kukula2010}.}
	\label{fig:hci}
 	\vspace{-0.3cm}
\end{figure}

The Human-Computer Interface (HCI) -- also known as the Human-Biometric System Interface (HBSI) -- is perhaps the most significant component of the entire system (Figure \ref{fig:hci}). A poorly conceived HBSI can result in the acquisition of poor quality data which, in turn, can exacerbate the errors of the biometric matcher. It can also undermine the usability and security of the entire system. A thoughtfully designed HBSI, on the other hand, will not only enhance user adoption of the technology~\cite{deluca2015} but also significantly improve user throughput in high volume applications such as border control systems. The HBSI will also play a pivotal role in procuring useful information when dealing with non-cooperative subjects in law enforcement applications. Human factors (and ergonomics) is a relatively understudied problem in the biometrics literature. 

{\bf \underline{Research problem \refstepcounter{custom}\thecustom}: Designing data-driven techniques for (a) modeling human behavior when interacting with a biometric system, and (b) using these models to redesign the human-computer interface for optimizing performance and enhancing usability.}

{\bf \underline{Research problem \refstepcounter{custom}\thecustom}: Developing energy-efficient multipurpose customizable biometric sensors that can not only acquire the biometric data of an individual but also rebuff adversarial attacks such as spoofing.}

{\bf \underline{Research problem \refstepcounter{custom}\thecustom}: Harnessing the principle of additive manufacturing for open-source physical production of innovative sensors and human-computer interfaces.}

\section{Smartphones and Wearable Devices}

\begin{figure}
\begin{center}
\includegraphics[width=3.2in]{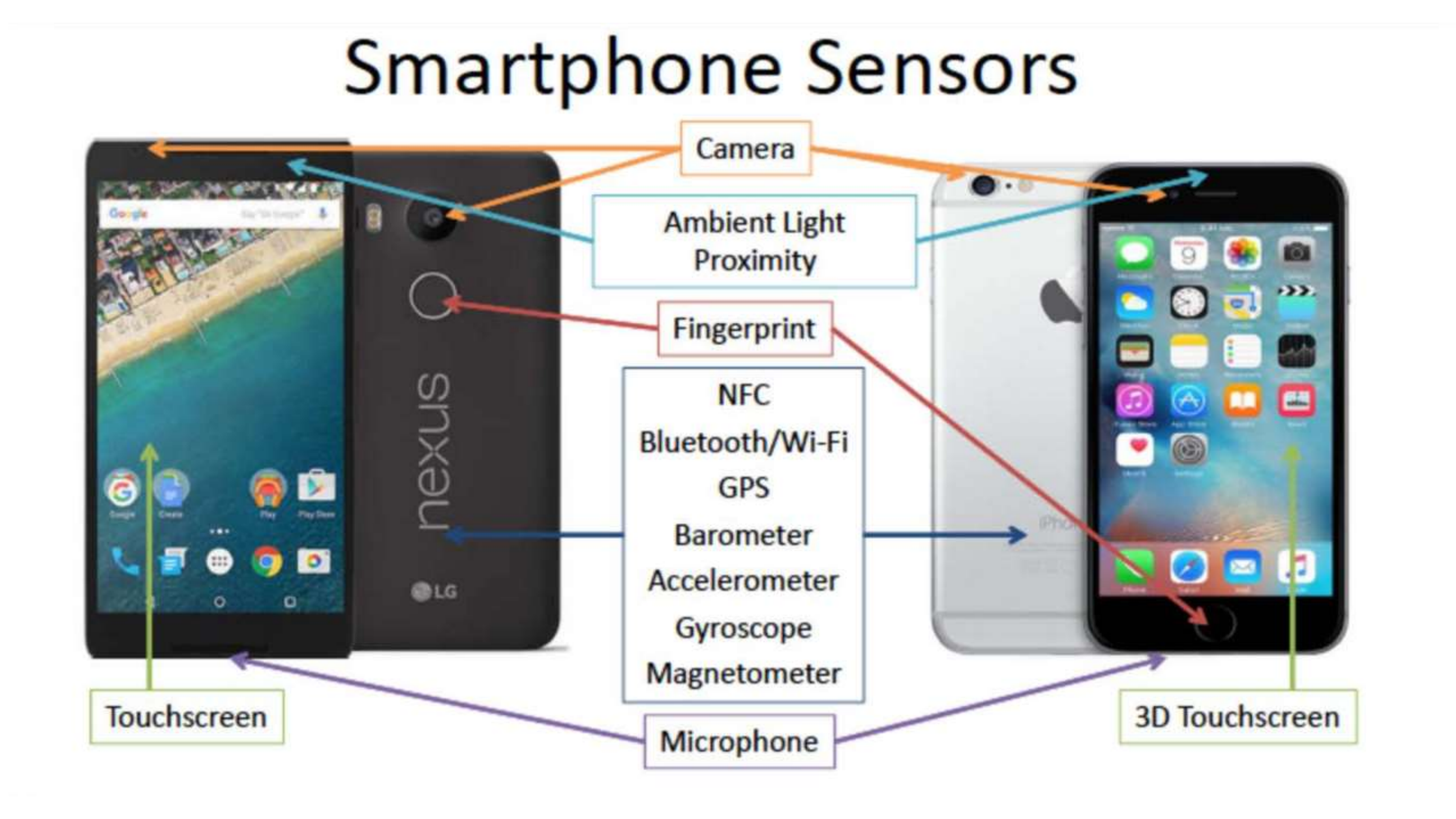}
\caption{Smartphones are equipped with a number of built-in sensors. Data from these non-biometric sensors can be aggregated to construct a behavioral signature of its owner. \textcopyright~Debayan Deb}
\label{fig:smartphone}
\end{center}
\vspace{-0.3cm}
\end{figure}

The explosive growth in the use of smartphones and wearable devices, such as smartwatches and activity trackers, presents an unprecedented opportunity for biometric researchers~\cite{Blasco2016,PatelContinuous2016}. Firstly, these devices store or transmit personal information (\eg, financial or health data), thereby requiring an effective mechanism to restrict access to the legitimate owner. Secondly, these devices are outfitted with a large number of sensors that record various physical (\eg, distance walked, posture) and biological (\eg, heart rate, skin temperature) attributes of an individual; principled methods are needed to parse through this heterogeneous data and distill a compact representation that can be used as the biometric signature of that individual (Figure \ref{fig:smartphone}). A related challenge is to be able to generate a ``portable" signature that can be used across devices belonging to the same individual.

{\bf \underline{Research problem \refstepcounter{custom}\thecustom}: Extracting a distinct and portable ``signature" of a subject from the data generated by the built-in sensors present in smartphones and wearable devices.}

{\bf \underline{Research problem \refstepcounter{custom}\thecustom}: Designing inexpensive biometric sensors for integration with smartphones and wearable devices.}

{\bf \underline{Research problem \refstepcounter{custom}\thecustom}: Developing computationally simple methods for active user authentication in resource-constrained wearable devices that have limited battery power.}

\section{Presentation Attack Detection}

\begin{figure}[h]
   
    \centering
    \includegraphics[width=0.4\textwidth]{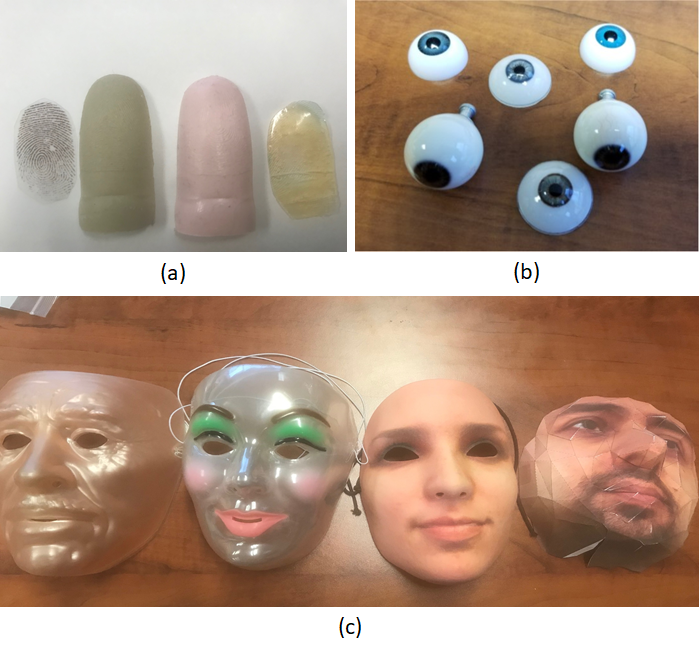}
    \caption{Examples of different types of spoof artifacts for (a) fingerprint, (b) iris and (c) face.}
    \label{fig:spoofs}
    
\end{figure}

\begin{figure}
\begin{center}
\includegraphics[width=3.2in]{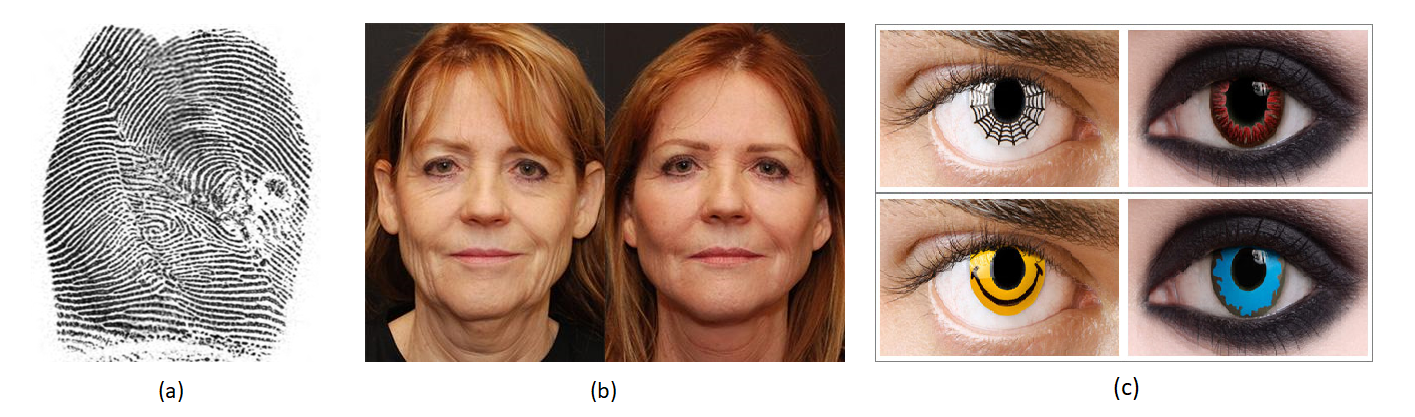}
\caption[Caption for LOF]{Examples of altered/obfuscated biometric traits. (a) A surgically altered fingerprint depicting transplantation with Z-cut. (b) A face image before and after plastic surgery.\protect\footnotemark~(c) Cosmetic contact lenses can obfuscate the underlying iris texture pattern.\protect\footnotemark}
\label{fig:altered}
\end{center}
\vspace{-0.3cm}
\end{figure}

A biometric system is vulnerable to presentation attacks where an adversary presents a fake (Figure \ref{fig:spoofs}) or altered (Figure \ref{fig:altered}) biometric trait to the sensor in order to fool the system~\cite{MarcelHBPAD2019}. Such an attack may be used to (a) enroll a fabricated trait in order to create a virtual identity that can be shared by a group of individuals; or (b) deliberately obfuscate one's own trait in order to evade being identified; or (c) spoof the biometric trait of another person in order to masquerade as them. A number of hardware-based and software-based solutions have been developed for presentation attack detection (PAD), especially for the face, fingerprint and iris modalities. However, most of the current solutions do not generalize well across different sensors and environments.

The challenge is to develop counter-measures that can deflect unseen and unknown attacks, \ie, those attacks that have not been considered as yet, but which the system will encounter in the future. This is a formidable challenge that can evolve into a ``cat-and-mouse" game between the adversary and the system designer. The advent of 3D printing and additive manufacturing will facilitate the generation of sophisticated presentation attack vectors. It is essential for PAD solutions to keep pace with these advanced technologies.          

{\bf \underline{Research problem \refstepcounter{custom}\thecustom}: Developing anti-spoofing methods that can generalize well across different types of attacks, sensors, environments, demographic groups and datasets.}


{\bf \underline{Research problem \refstepcounter{custom}\thecustom}: Designing presentation attack detectors for resource-constrained IoT devices such as smartphones and wearable devices.}

{\bf \underline{Research problem \refstepcounter{custom}\thecustom}: Automating the process of generating adversarial spoof artifacts for a specific biometric sensor by combining 3D printing technology with robotic process automation testing.}

\setcounter{footnote}{1}  

\setcounter{footnote}{0}

\begin{table*}[tbph]
    \centering
    \footnotesize
    \caption{Current state-of-the-art performance of four biometric modalities. Verification (V) performance is reported using the False Match Rate (FMR) and False Non-Match Rate (FNMR). Identification (I) performance is reported using the False Negative Identification Rate (FNIR) and False Positive Identification Rate (FPIR).}
    \begin{tabular}{||l|c|l|l|l|l|l||}
        \hline
        \multirow{2}{*}{\bf Modality} & \multirow{2}{*}{\bf Mode}    & \multirow{2}{*}{\bf Report} & {\bf Template} & \multirow{2}{*}{\bf Performance} & \multirow{2}{*}{\bf Dataset Size} & \multirow{2}{*}{\bf Comments}\\
         & & & {\bf  Size} & & & \\\hline \hline
        \multirow{2}{*}{Face} & V & FRVT 1:1~\cite{Grother2019-FRVT11_Apr19} & 1.4 KB & FNMR 2.71\% @ FMR 0.01\% & 1K subjects, 100K images & ``In-the-wild" \\\cline{2-7}
                              & I & FRVT  1:N~\cite{Grother2019-FRVT1N_Nov18} & 2.05 KB & FNIR 2.0\% @ FPIR 0.1\% & 485K Probe, 1.6M Gallery & Mugshots\\\hline
        \multirow{2}{*}{Fingerprint} & V & FVC-onGoing\protect\footnotemark & 5.9 KB & FNMR 0.036\% @ FMR 0.01\% & 115,710 comparisons & Operational conditions \\\cline{2-7}
                                 & I & FpVTE~\cite{Watson2015-FpVTE} & 6.1 KB & FNIR 1.9\% @ FPIR 0.1\% & 30K Probe, 100K Gallery & Plain fingerprint \\\hline
        \multirow{2}{*}{Iris} & V & \multirow{2}{*}{IREX IX~\cite{quinn2018irex}} & 12.33 KB & FNMR of 0.57\% @ FMR 0.001\% &260,809 subjects, & Both-eyes;\\\cline{2-2}\cline{4-5}
                              & I & & 12.33 KB & FNIR of 0.67\% @ FPIR 0.1\% & 673,662 samples & High-quality\\\hline
        Voice & V & SRE 2016 \cite{sadjadi20172016} & \textcolor{gray}{not given}& FNMR of 39\% @ FMR 0.01\% & 201 speakers & Multi-lingual \\\hline
    \end{tabular}
    \label{tab:performance}
\end{table*}

\section{Poor Quality Data}
\begin{figure}
    \centering
    \hspace{0.1cm}
    \includegraphics[height=3.75cm]{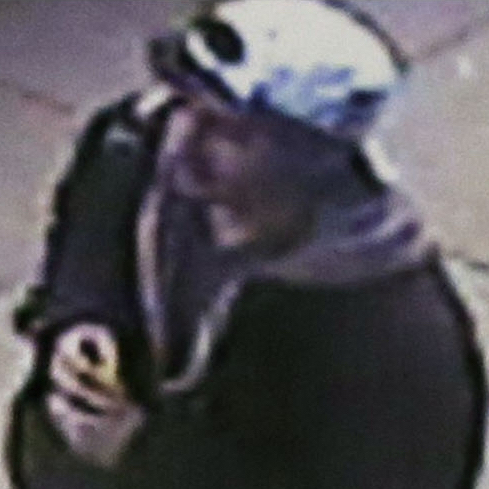}
    \hfill
    \includegraphics[height=3.75cm]{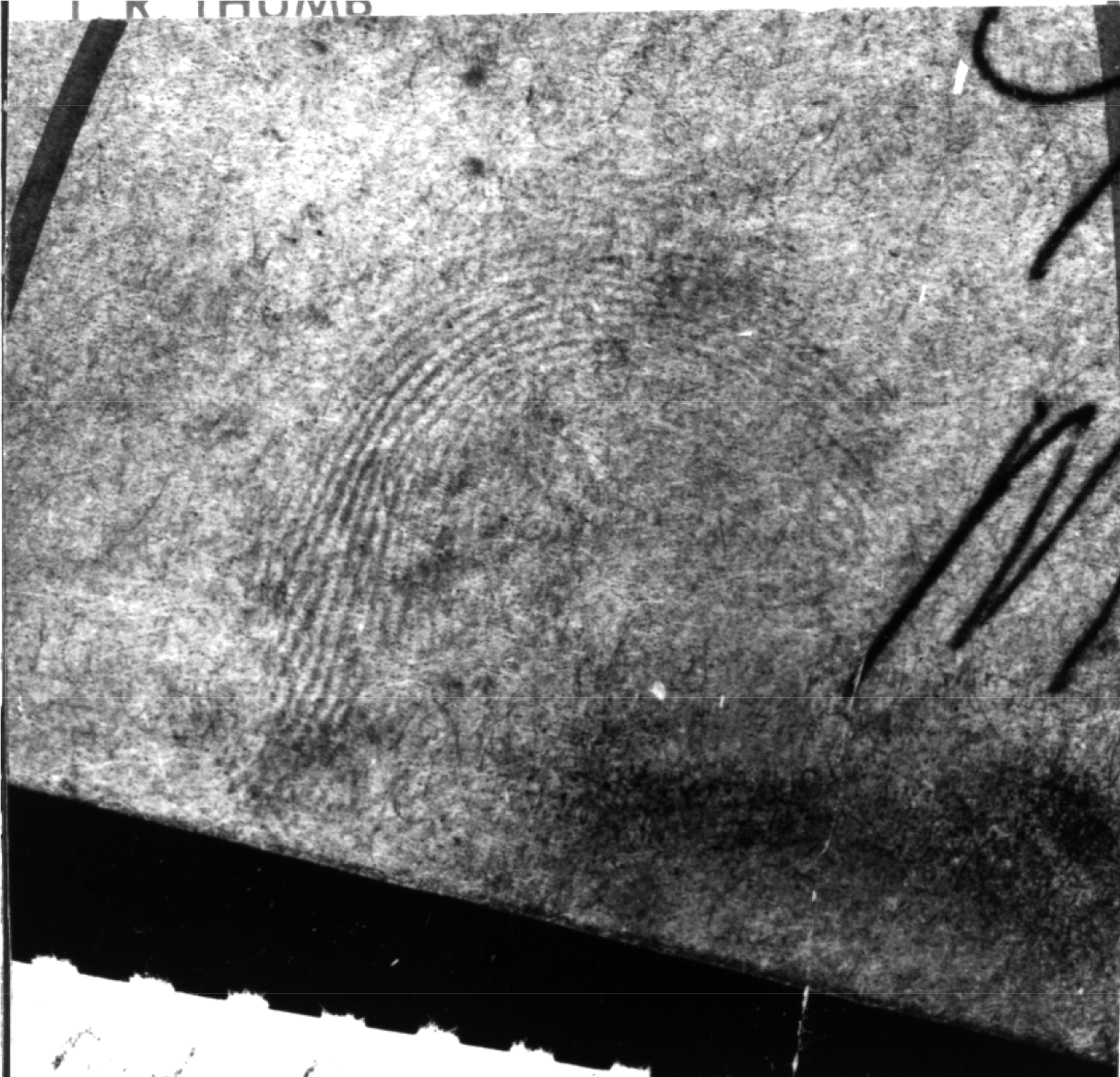}
    \hspace{0.1cm}
    \caption[Caption for LOF]{Examples of poor quality data: face image of the Boston bomber captured using a surveillance camera (left) and a latent fingerprint on the surface of an object (right).}
    \label{fig:poorquality}
    \vspace{-0.5cm}
\end{figure}

As can be seen in Table \ref{tab:performance}, state-of-the-art biometric matchers exhibit very good performance when the quality of the input data is reasonably good. However, the performance sharply degrades when a matcher encounters poor quality data~\cite{BharadwajQuality2014}. Examples of poor quality data include fingermarks lifted from a crime scene, audio data recorded in noisy environment, iris images awash with strong non-uniform illumination, or partially occluded low-resolution faces in surveillance videos (Figure \ref{fig:poorquality}). Reliably enhancing such data is a challenging problem since many data enhancement algorithms do not explicitly attempt to preserve the {\em biometric content} of the input data; this can potentially alter the biometric cue in the data resulting in its inadvertent match with an incorrect identity. This can have serious consequences in law enforcement applications. Further, when using poor quality biometric data as evidence in a court-of-law, it is necessary to first compute its {\em evidential value}, \ie, its utility for reliably identifying an individual.  

{\bf \underline{Research problem \refstepcounter{custom}\thecustom}: Developing methods for enhancing poor quality data such that the biometric content is not unduly perturbed.}

{\bf \underline{Research problem \refstepcounter{custom}\thecustom}: Designing robust feature extraction and matching algorithms that can successfully operate on poor quality data.}

{\bf \underline{Research problem \refstepcounter{custom}\thecustom}: Computing the evidential value of poor quality biometric data, especially in forensic applications.}

\addtocounter{footnote}{0} 
\footnotetext{\url{https://biolab.csr.unibo.it/fvcongoing/}}
\stepcounter{footnote}\footnotetext{\url{https://www.cincyfacialplastics.com/before-and-after/luxe-lift-pictures-5859}}
\stepcounter{footnote}\footnotetext{\url{https://hoovervisioncenter.com/2015/10/21/halloween-hazard-the-dangers-of-cosmetic-contact-lenses/}}

\section{Data Integrity}

\begin{figure}[tbhp]
   
    \centering
    \includegraphics[width=0.3\linewidth]{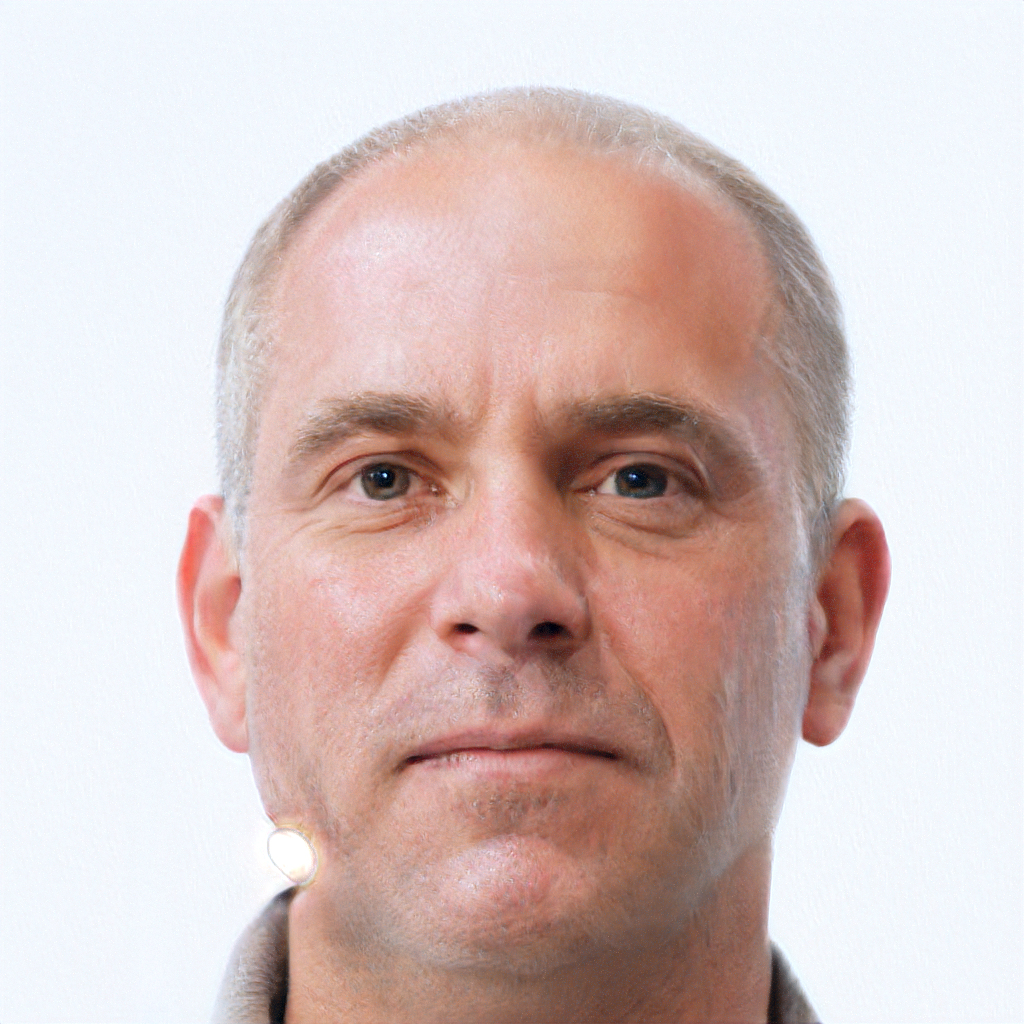}
    ~
    \includegraphics[width=0.3\linewidth]{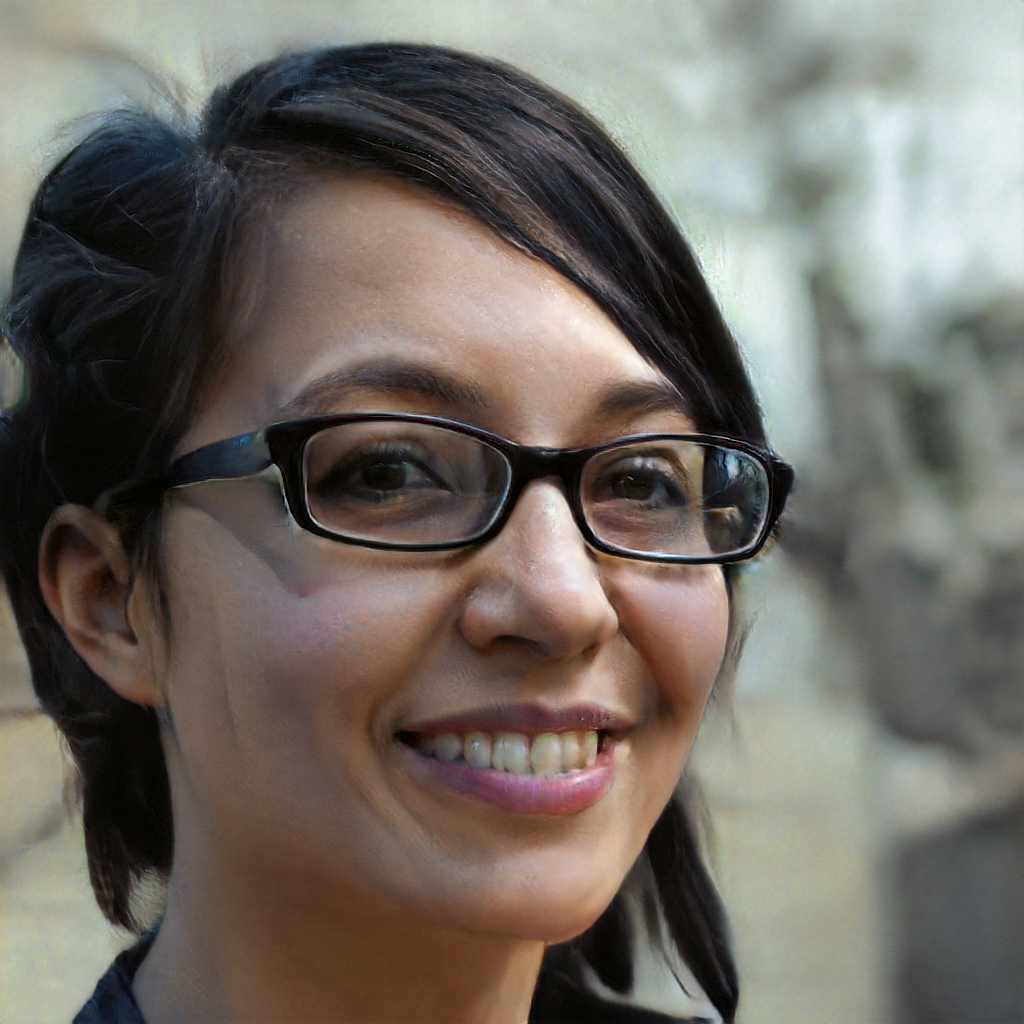}
    ~
    \includegraphics[width=0.3\linewidth]{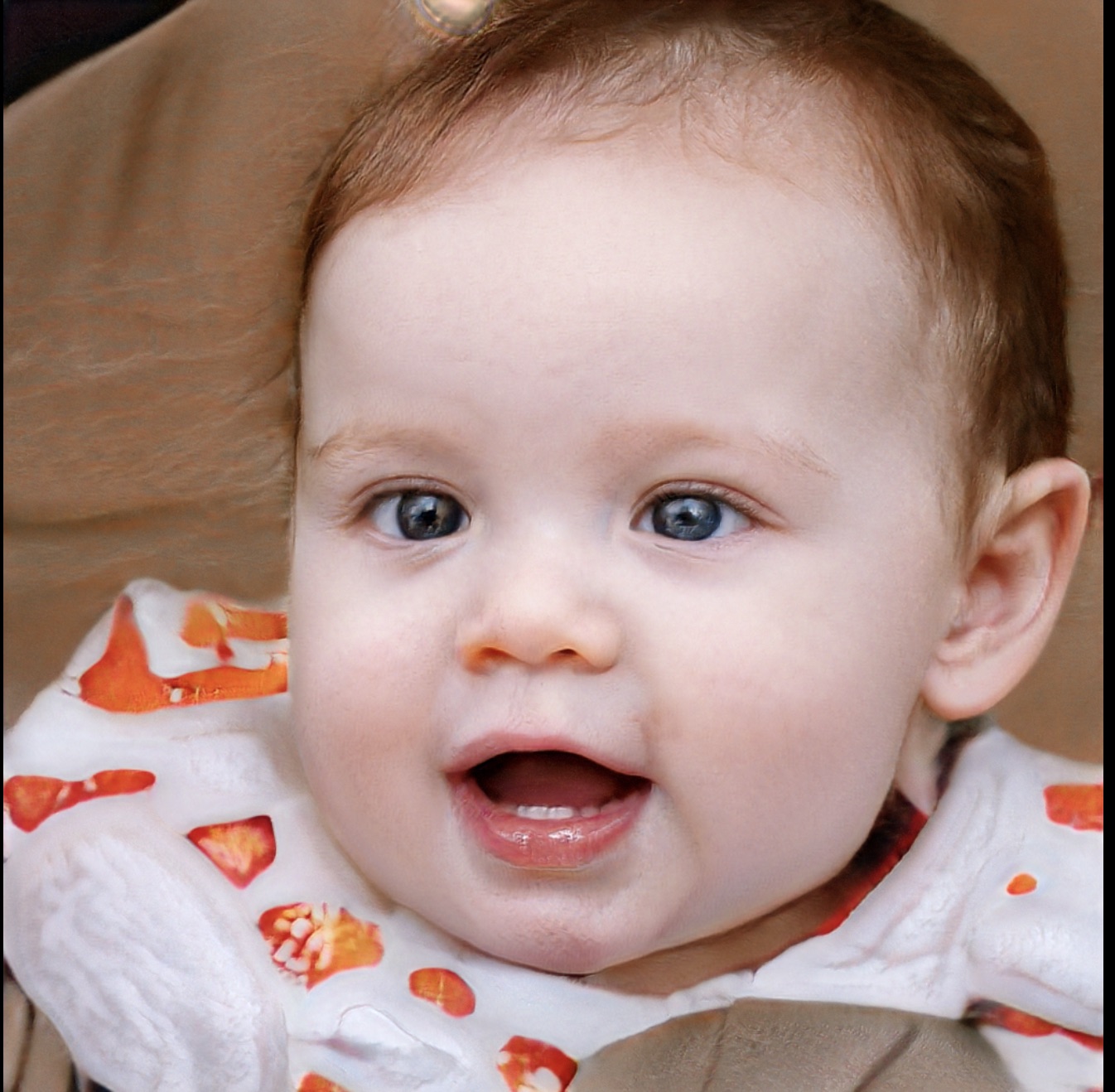}
    \caption{Examples of realistic-looking digital face images generated using GANs. Taken from {\em thispersondoesnotexist.com}.}
    \label{Fig:GANface}

\end{figure}

The {\em integrity} of the raw biometric data (\eg, face image) is of paramount importance, especially when it is used as evidence in a court of law. However, the raw biometric data of a person can be maliciously modified for nefarious purposes. For example, digital images and videos of a person's face may be subtly modified using an editing tool such as Adobe Photoshop with the intention of creating a false match. The problem is compounded when strategically altered images and videos are displayed on the Web, particularly on image sharing platforms. This leads to a proliferation of doctored images and their duplicates on the Web, making it difficult to identify the original (pre-modified) image. Recently, Generative Adversarial Networks (GANs) have been used to synthesize \textit{realistic} looking images and videos known as {\em DeepFakes} (Figure ~\ref{Fig:GANface}).

\begin{figure}[t]
   
    \centering
    \includegraphics[width=\linewidth]{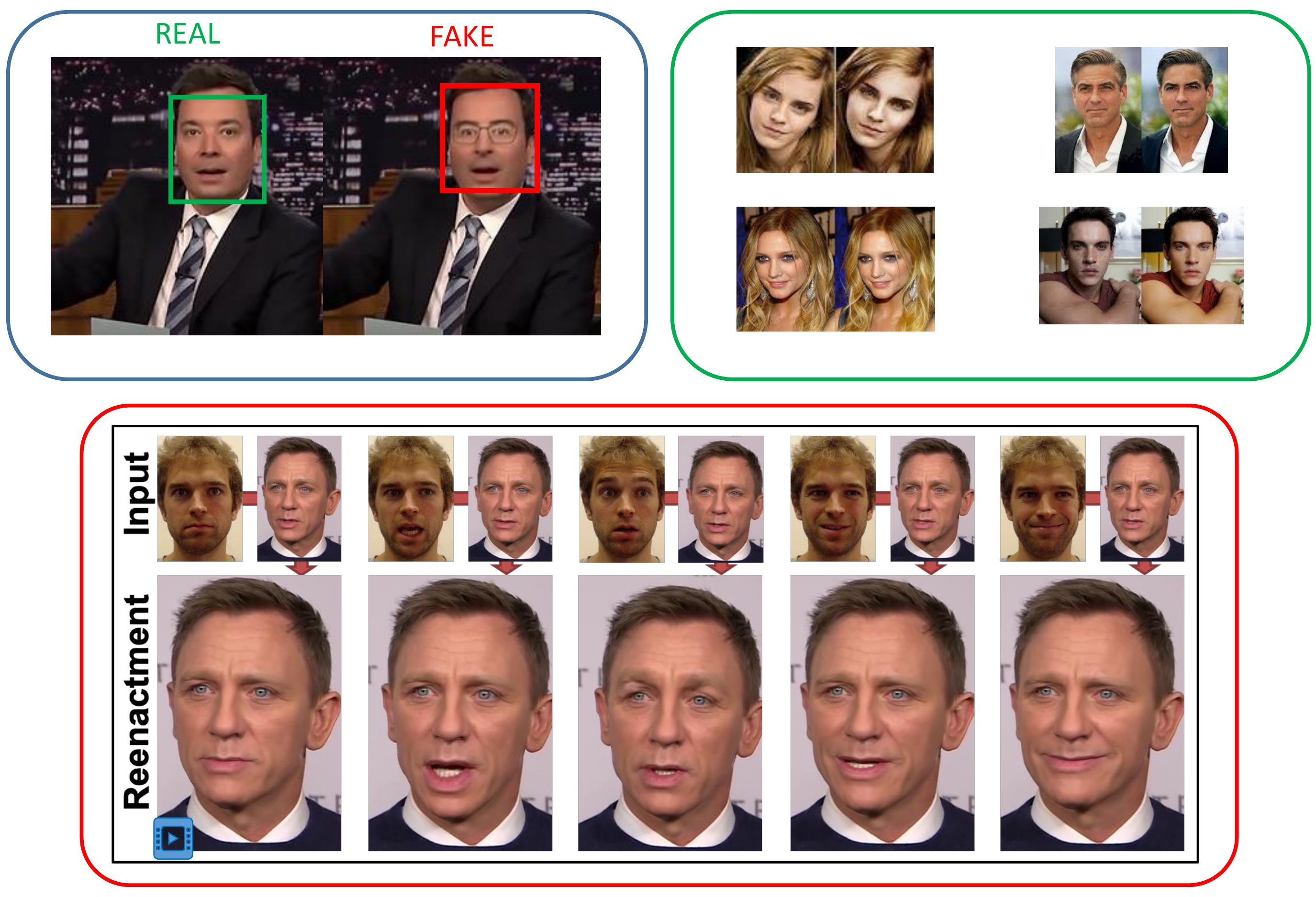}
    \caption[Caption for LOC]{Examples of digitally modified media. Top left: A GAN-generated face image.\protect\footnotemark~Top right: Near duplicate face images of four subjects created using Photoshop.\protect\footnotemark~Bottom: Rendering of a ``fake" video by transferring the facial expressions of one individual (source) to another individual (target).\protect\footnotemark}
    \label{Fig:Image_For}

\end{figure}

When images posted on the Web are used for determining the identity of a person, an erroneous match due to perturbations in the images can have serious consequences. Therefore, it is essential for a biometric system to validate the integrity of the input digital media prior to processing it (Figure~\ref{Fig:Image_For}). The principle of digital image forensics~\cite{ImageForensics_Farid} can be used for this purpose.

{\bf \underline{Research problem \refstepcounter{custom}\thecustom}: Designing algorithms for detecting {\em DeepFakes} as well as maliciously modified images/videos of a person.}

{\bf \underline{Research problem \refstepcounter{custom}\thecustom}: Developing methods to identify the original unmodified image from a given set of near-duplicate biometric images and to infer the trail of photometric and geometric modifications that produced the other images.}

{\bf \underline{Research problem \refstepcounter{custom}\thecustom}: Designing robust sensor-fingerprinting algorithms that can link a biometric image to the specific sensor unit that produced it.}

 \addtocounter{footnote}{-3} 

\stepcounter{footnote}\footnotetext{\url{https://www.deeptracelabs.com/}}

\stepcounter{footnote}\footnotetext{\url{https://www.buzzfeed.com/jessicamisener/23-celebrities-before-after-photoshop}}

\stepcounter{footnote}\footnotetext{\url{https://web.stanford.edu/~zollhoef/papers/CVPR2016_Face2Face/page.html}}

\section{Cross-modal Biometrics}

\begin{figure}[t]		
	\centering
	\includegraphics[width=0.9\linewidth]{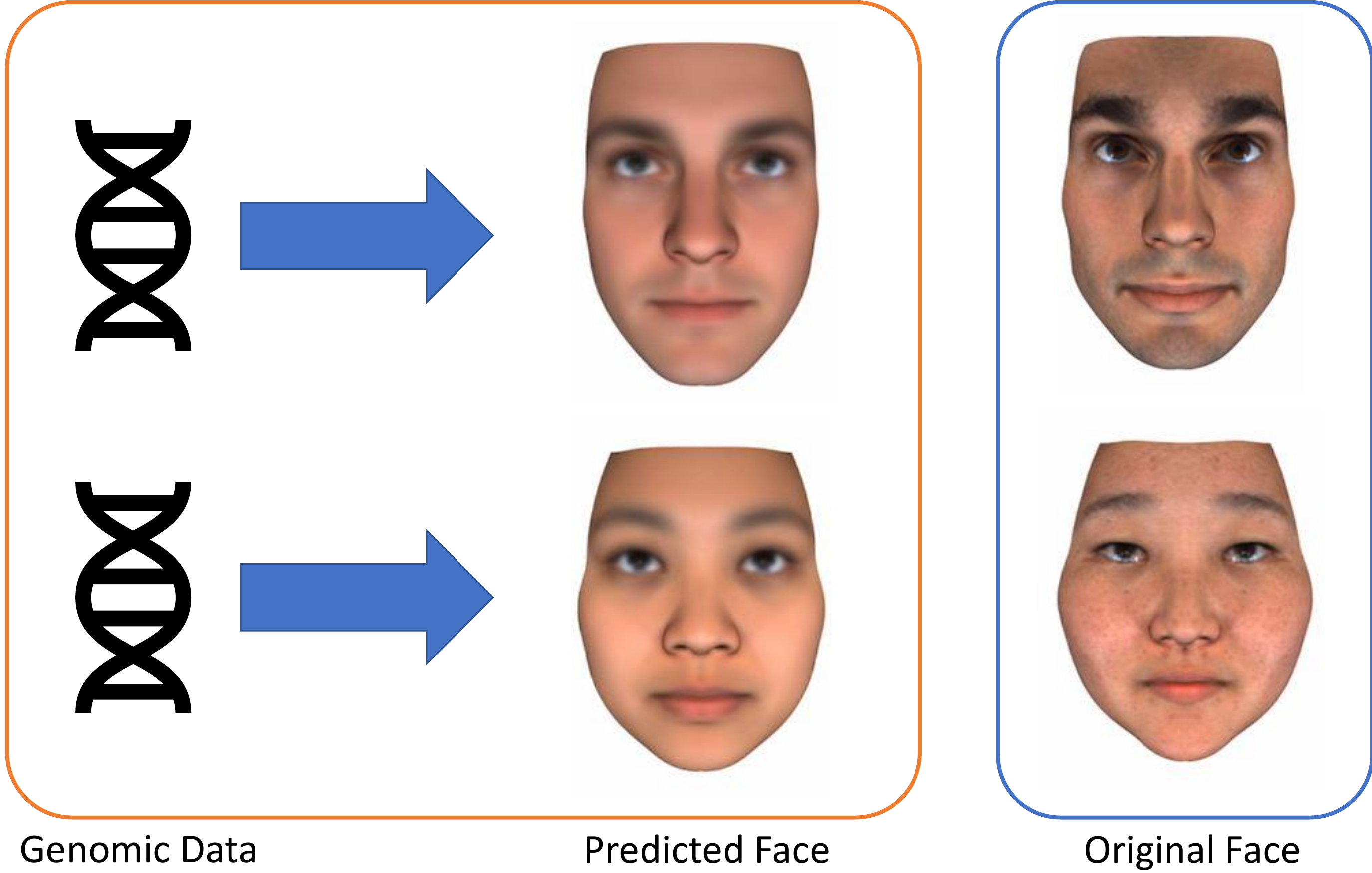}
	\caption{Illustration of cross-modal biometrics where a face image is generated from the DNA sample of a subject. Adapted from Lippert \etal~\cite{lippert2017identification}.}
	\label{fig:dnatoface}
 	\vspace{-0.3cm}
\end{figure}

{\em Cross-modal} matching involves associating the data pertaining to one biometric modality with that of another modality. There is limited work on this topic. One application where cross-modal association can be beneficial is in the mapping of genomic data to phenotypic traits~\cite{lippert2017identification}. For example, generating the face image of a person from their DNA sample can be useful in criminal investigations (Figure~\ref{fig:dnatoface}). Cross-modal matching can also be useful in applications where a specific biometric trait may not be consistently available. For example, in the case of an indoor surveillance video, the face image of a subject may not be available in every frame due to low image-resolution, poor illumination, non-frontal pose or occlusions. However, the audio of the subject's voice may be available in such frames. The ability to associate the voice samples of a subject with their face images for cross-modal identity matching can be valuable. The matching itself can be accomplished by the extraction of common soft biometric cues from the two modalities (\eg, age, gender, height, weight) or by modeling the correlation between the morphological aspects of a subject's face and the acoustic characteristics of their voice~\cite{nagrani2018seeing}. Another example of cross-modal matching would be linking RGB face images in a face database with near-infrared (NIR) iris images in an iris database~\cite{jillela2014matching}.

\begin{figure}[htpb]		
	\centering
	\includegraphics[scale=0.7]{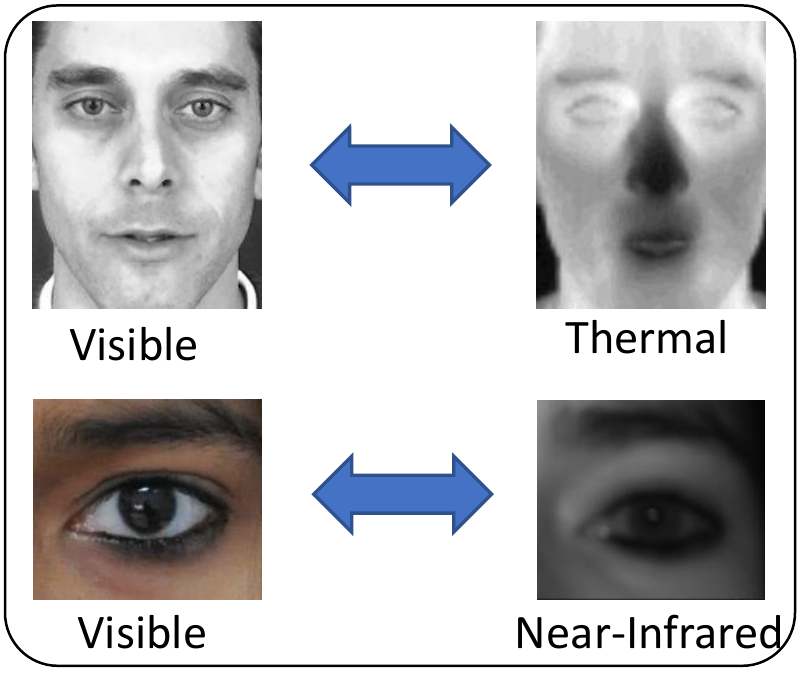}
	\caption{Heterogeneous biometric recognition in the context of face (top) and iris (bottom).}
	\label{fig:heterogeneous-face}
 	\vspace{-0.3cm}
\end{figure}

A related problem is {\em heterogeneous} matching that involves comparing data originating from two distinctly different sensors but pertaining to the same modality, \eg, cross-spectral matching of RGB face images with thermal face images (Figure \ref{fig:heterogeneous-face}). While it may not be as challenging as cross-modal biometrics, it nevertheless is an unsolved problem notwithstanding the large number of papers on the topic. 

{\bf \underline{Research problem \refstepcounter{custom}\thecustom}: Developing methods for establishing the degree of correlation, at the biological level, between two or more biometric traits.}

{\bf \underline{Research problem \refstepcounter{custom}\thecustom}: Developing techniques to deduce phenotypic attributes from genomic data for cross-modal biometric matching.}

{\bf \underline{Research problem \refstepcounter{custom}\thecustom}: Designing methods for generating a canonical representation of one biometric trait from another trait.}

{\bf \underline{Research problem \refstepcounter{custom}\thecustom}: Developing models to assess the upper bound on the recognition accuracy of cross-spectral biometric recognition.}

\section{Postmortem Biometrics}

In some scenarios, it may be necessary to identify a deceased person by comparing their postmortem (PM) biometric data with the corresponding antemortem (AM) data. For example, police officers have sought to unlock the smartphone of a deceased suspect by using the suspect's postmortem fingerprint.\footnote{\url{https://www.livescience.com/62393-dead-fingerprint-unlock-phone.html}} Another application is victim identification using postmortem data in the wake of mass fatalities due to tsunamis, terrorist attacks, wars, earthquakes or nuclear explosions. While dental radiographs have been traditionally used for postmortem biometric identification (especially in the context of mass disasters), there is increasing interest in utilizing other biometric modalities, such as face, fingerprints and iris, for this purpose~\cite{Trokielewicz2019}. However, postmortem biometric identification is beset with a number of challenges due to reduced data quality and natural decomposition of body parts. Furthermore, in many cases, the feature extraction and matching algorithms developed for antemortem data may not be able to successfully process postmortem data. Factors such as cause of death, subject's age, environmental conditions, \etc, can also impact the biometric utility of certain traits.

{\bf \underline{Research problem \refstepcounter{custom}\thecustom}: Designing novel quality enhancement, segmentation and feature extraction algorithms for processing postmortem biometric data.}

{\bf \underline{Research problem \refstepcounter{custom}\thecustom}: Developing methods to effectively match postmortem data of a subject with the corresponding antemortem data.}

{\bf \underline{Research problem \refstepcounter{custom}\thecustom}: Modeling the temporal degradation in postmortem biometric data and determining the factors that impact the biometric utility of postmortem data.}

{\bf \underline{Research problem \refstepcounter{custom}\thecustom}: Investigating the feasibility of using postmortem biometric traits to launch a presentation attack against a biometric system.}

\section{Soft Biometrics}

While biometric data is typically used for {\em recognizing} an individual, it is possible to deduce or extract additional information, such as age, gender, ethnicity, height, weight, hair color, eye color, clothing style, tattoos, \etc, from the same data~\cite{dantcheva_what_2016}. These attributes, sometimes referred to as soft biometric attributes, can be used independently or in conjunction with primary biometric traits to improve the recognition accuracy of a biometric system. Further, they provide a human-interpretable description of the underlying biometric data (\eg, ``Young Asian Male" or ``Iris With a Dilated Pupil"). They can also be used to restrict the search space in a gallery database to only those identities that share similar soft biometric attributes as the input probe data. In spite of their potential benefits, there are a number of open research problems in this area.       

{\bf \underline{Research problem \refstepcounter{custom}\thecustom}: Designing methods to deduce soft biometric information from poor quality biometric data.}

{\bf \underline{Research problem \refstepcounter{custom}\thecustom}: Extracting soft biometric information from behavioral traits.}

{\bf \underline{Research problem \refstepcounter{custom}\thecustom}: Estimating upper bounds on the uniqueness and persistence of soft biometric attributes.}

\section{Personal Privacy}


As stated in the previous section, recent advances in machine learning has made it possible to extract ancillary information, such as a person's age, gender, ancestral origin and health, from their biometric data using sophisticated classifiers~\cite{dantcheva_what_2016}. The possibility of eliciting genetic information from facial images has also been studied~\cite{Martinez2018}. When such information is extracted without the subject's consent, then issues of {\em function creep} and {\em privacy infringement} are brought to the fore. Similarly, when biometric data is used to unmask the identity of a person by {\em linking} information from seemingly disparate sources, it can represent a privacy breach. For example, matching an {\em unidentified} face image from a pseudonymized dating website with an {\em identified} face image in a social network website can expose sensitive details about a person through data accretion~\cite{acquisti_face_2014}. 

\begin{figure}[t]		
	\centering
	\includegraphics[width=0.9\linewidth]{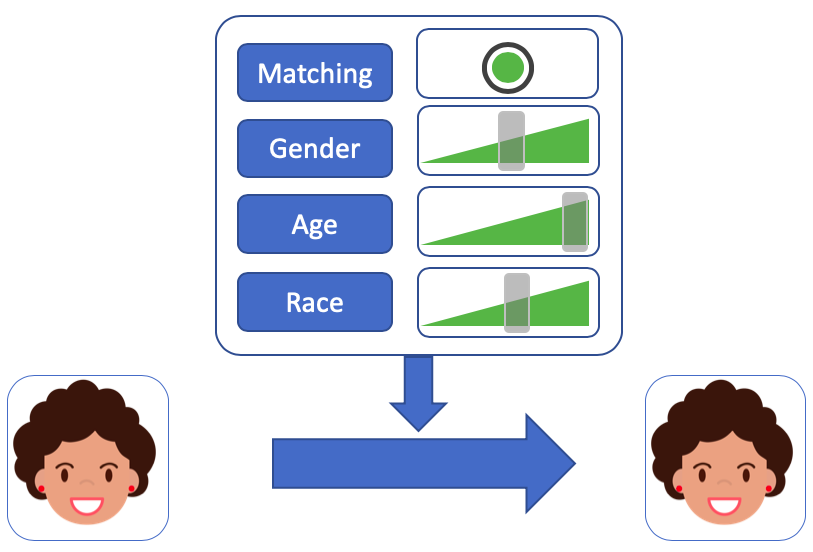}
	\caption{Illustration of controllable biometric privacy, where users can determine which information to keep and which to conceal.}
	\label{fig:controllable-privacy}
 	\vspace{-0.3cm}
\end{figure}

Recent research has explored the notion of controllable privacy~\cite{sim_controllable_2015} where specific ancillary cues are suppressed in the raw image (Figure \ref{fig:controllable-privacy}). For example, semi-adversarial neural networks have been designed to remove gender cues from a face image, through a series of perturbations, such that the performance of automated gender classifiers is confounded but the performance of face matchers is retained~\cite{mirjalili_gender_2018}. Introduction of the EU General Data Protection Regulation (GDPR) has reinforced the importance of designing privacy-preserving methods in the context of biometric systems.  

{\bf \underline{Research problem \refstepcounter{custom}\thecustom}: Designing methods for imparting controllable and quantifiable soft-biometric privacy to biometric data without compromising recognition accuracy.}

\section{Biases in Biometrics}

\begin{figure}[htpb]		
	\centering
	\includegraphics[width=0.8\linewidth]{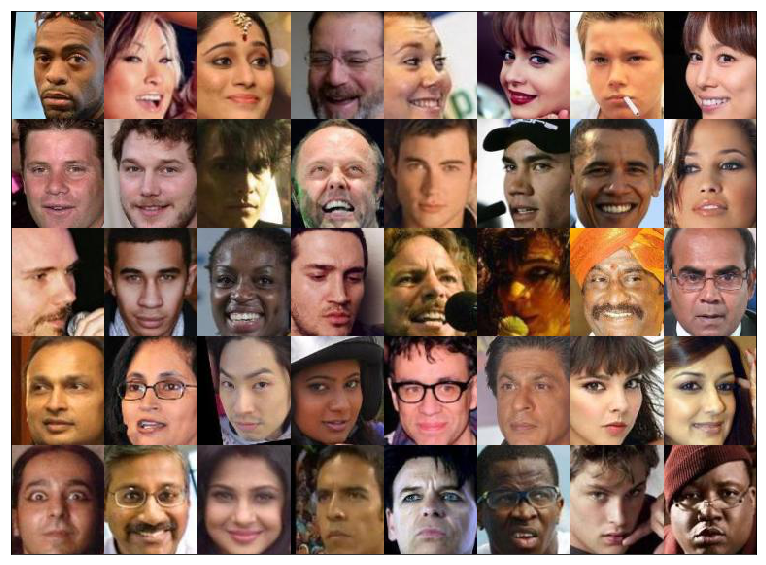}
	\caption{It is essential for biometric systems to be unbiased and perform well across diverse demographic groups (images are from the CelebA dataset).}
	\label{fig:unbias}
 	\vspace{-0.3cm}
\end{figure}

Biometric systems, especially those based on face recognition, have exhibited demographic {\em bias} in which certain population groups have experienced significantly higher error rates than others. For example, face detection methods have been observed to fail more often on subjects with darker skin-tone than those with lighter skin-tone~\cite{buolamwini_gender_2017}. In another study, researchers found that automated face recognition systems developed in Western countries performed better on Caucasian faces than East Asian faces and, conversely, automated face recognition systems developed in East Asian countries performed better on East Asian faces than Caucasian faces~\cite{Phillips2011}.  While such biases could be attributed to the lack of sufficiently diverse training data, it nevertheless brings into question the fairness and integrity of AI-based systems. Indeed, data-driven approaches seem to be vulnerable to such biases and it remains to be seen how this can be mitigated in the context of biometric systems that are increasingly being deployed in heterogeneous populations worldwide (Figure \ref{fig:unbias}). 

{\bf \underline{Research problem \refstepcounter{custom}\thecustom}: Determining the underlying cause for race and gender bias in biometrics and to design methods that alleviate this problem.}

{\bf \underline{Research problem \refstepcounter{custom}\thecustom}: Assembling large multimodal biometric datasets exhibiting demographic diversity in order to effectively train biometric matchers.}

\section{Other Research Problems}

Besides the aforementioned topics, there are a number of other active research problems in the field. These include: (a) designing novel sensors for acquiring biometric data from infants and toddlers; (b) performing biometric recognition based on bacterial colonies on the human skin; (c) reliable human recognition from low-quality contaminated DNA samples; (d) homomorphic encryption methods for biometric template security; (e) integration of biometrics in the blockchain protocol for implementing self-sovereign identity; (f) information fusion techniques for combining biometrics with demographic data, quality measures, social information, and presentation attack detectors; (g) studying the impact of age and disease on the performance of individual biometric traits; (h) discovering and mitigating the impact of adversarial samples that can destabilize Deep Learning based biometric matchers; (i) harnessing explainable AI techniques for semantically interpreting trained neural network models; and (j) models for predicting biometric performance of large-scale systems having billions of identities.

\section{Summary}
Biometrics is a fascinating pattern recognition problem with several societal benefits~\cite{Pato2010}. The past decade has seen a surge in the use of biometric technology for diverse applications. Advancements in other domains have opened up new opportunities for biometric researchers. At the same time, a number of fundamental issues remain unsolved in the field even after several years~\cite{Jain2004Grand}. In this paper, we highlighted some of the research opportunities in biometrics and discussed its intersection with adjacent fields including forensics, genomics, anthropology and psychology.

The definition of {\em identity} itself is continually evolving~\cite{Bostrom2011-Identity}. In an increasingly connected world,\footnote{As of April 2019, $\sim$55\% of the world's population has internet access and there are $\sim$25 billion IoT devices.} the distinction between {\em social identity}, {\em online identity} and {\em device identity} has blurred. Individuals are increasingly leaving their ``digital fingerprints" on the Web and in personal electronic devices such as smartphones and wearables. This, coupled with the widespread availability of inexpensive digital sensors and storage units, has led to the realization of {\em exoself}, where the physical identity of a person overlaps significantly with their digital and device identities. This has further enhanced the scope of biometrics thereby bringing together evidence at the molecular level, biological level, behavioral level and digital level for human recognition.

{\footnotesize
\bibliographystyle{ieee}
\bibliography{refs}
}

\end{document}